\documentclass[letterpaper, 10 pt, journal, twoside]{ieeetran}

\IEEEoverridecommandlockouts
\usepackage{graphics}
\usepackage{epsfig}
\usepackage{mathptmx}
\usepackage{times}
\usepackage{amsmath}
\usepackage{amssymb}

\usepackage[hidelinks]{hyperref}

\usepackage{algorithm}
\usepackage{algpseudocode}

\DeclareMathAlphabet{\mathcal}{OMS}{cmsy}{m}{n}
\DeclareMathOperator*{\argmax}{arg\,max}

\newcommand\Tau{\mathcal{T}}

\usepackage[font=small,skip=3pt]{caption}
\usepackage{subcaption}

\usepackage[dvipsnames]{xcolor}

\author{Siddhant Gangapurwala, Alexander Mitchell and Ioannis Havoutis%
\thanks{Manuscript received: September, 10, 2019; Revised December, 17, 2019;
Accepted January, 19, 2020.}
\thanks{This paper was recommended for publication by Editor Nikos Tsagarakis
upon evaluation of the Associate Editor and Reviewers' comments.
This work was supported by the UKRI/EPSRC RAIN and ORCA Hubs
[EP/R026084/1, EP/R026173/1], Robust Legged Locomotion [EP/S002383/1] and the
EU H2020 Projects MEMMO and THING. It was conducted as part of ANYmal Research,
a community to advance legged robotics.}
\thanks{The authors are with the Dynamic Robots Systems Group, Oxford Robotics
Institute, University of Oxford, UK. Email:
{\tt\small \{siddhant,ioannis\}@robots.ox.ac.uk}.}}

\title{Guided Constrained Policy Optimization for\\Dynamic Quadrupedal Robot
Locomotion}

\begin{document}

\maketitle

%% Abstract
\begin{abstract}
Deep reinforcement learning (RL) uses model-free techniques to optimize
task-specific control policies. Despite having emerged as a promising approach
for complex problems, RL is still hard to use reliably for real-world
applications. Apart from challenges such as precise reward function tuning,
inaccurate sensing and actuation, and non-deterministic response, existing RL
methods do not guarantee behavior within required safety constraints that are
crucial for real robot scenarios. In this regard, we introduce guided
constrained policy optimization (GCPO), an RL framework based upon our
implementation of constrained proximal policy optimization (CPPO) for tracking
base velocity commands while following the defined constraints. We introduce
schemes which encourage state recovery into constrained regions in case of
constraint violations. We present experimental results of our training method
and test it on the real ANYmal quadruped robot. We compare our approach
against the unconstrained RL method and show that guided constrained RL offers
faster convergence close to the desired optimum resulting in an optimal, yet
physically feasible, robotic control behavior without the need for precise
reward function tuning.

\end{abstract}

%% Keywords

%% Introduction
\section{Introduction}
\label{introduction}
\IEEEPARstart{L}{egged} locomotion has been an active area of robotics
research over the past few decades. Despite our best efforts, achieving
extraordinarily dynamic robotic behavior still remains an open problem. Most
of the existing work has focused on the use of traditional model-based control
techniques, such as offline trajectory optimization
(TO)~\cite{betts1998survey} and online model predictive control
(MPC)~\cite{neunert2018whole} which, due to their mathematical complexity, are
often based on simplified models of the systems. Such simplifications result
in control solutions that are often mechanically limiting and inefficient.

Considering robotic locomotion as a reinforcement learning (RL) problem
\cite{sutton2018reinforcement} offers a model-free data-driven alternative to
model-based control. Although RL has witnessed significant contributions from
researchers to address issues such as sample
inefficiency~\cite{pietquin2011sample} and hyperparameter
tuning~\cite{schulman2017proximal}, it still faces significant challenges to
be used for real-world robotic locomotion applications mainly due to no hard
guarantees on safety-critical constraints.

In this work, we develop an RL problem formulation that introduces constraints
based on optimal control techniques. We train a control policy, using
our constrained proximal policy optimization (CPPO) method based upon proximal
policy optimization (PPO)~\cite{schulman2017proximal}, for tracking
user-generated reference base velocities on the ANYmal~\cite{hutter2016anymal}
quadruped, a 33 kg legged robot. We experimentally validate its
performance in comparison with unconstrained training procedures in a
physically realistic simulation environment and on the real ANYmal robot.

\begin{figure}[tb]
  \centering
  \includegraphics[width=0.45\textwidth]{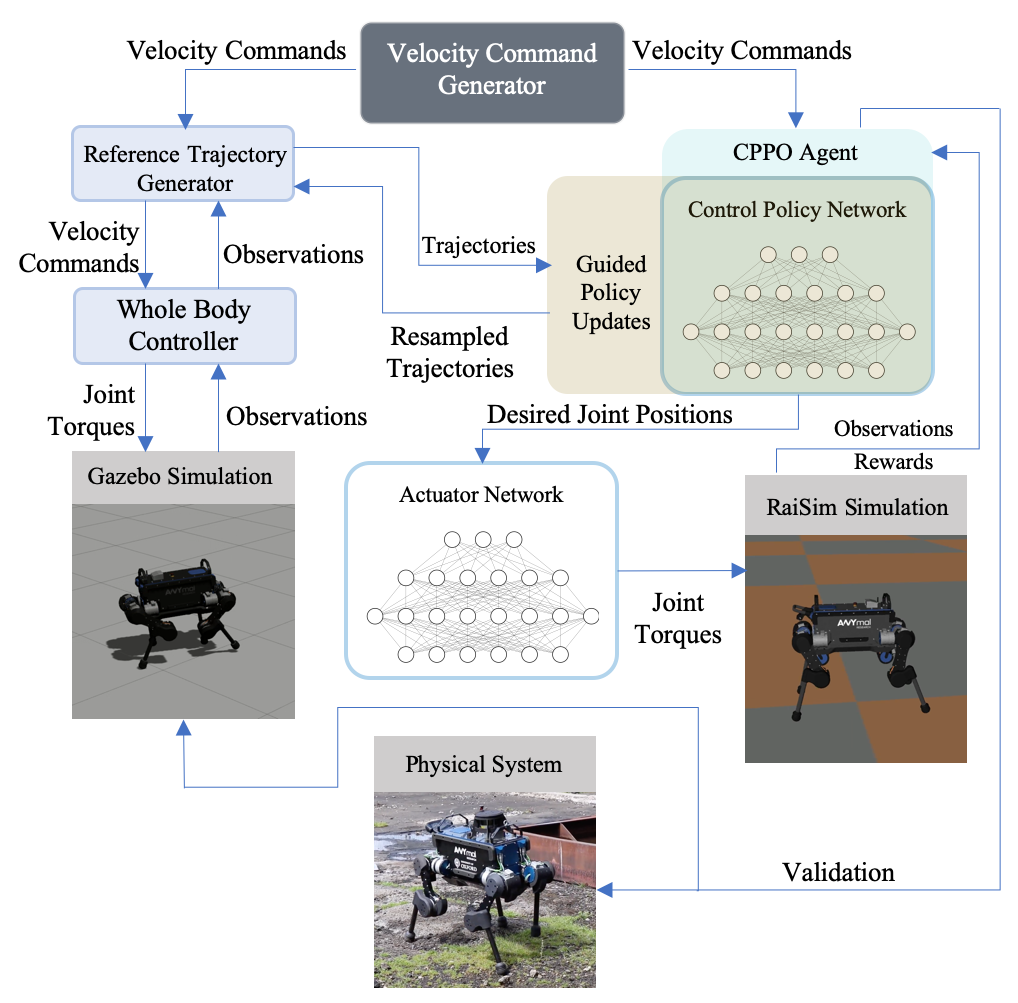}
  \caption{Overview of our training and validation process.
  Accompanying video can be found at:
  \href{https://youtu.be/iPDmG9knkLs}{\footnotesize
  https://youtu.be/iPDmG9knkLs}}
  \label{fig:overview}
\end{figure}

\subsection{Related Work}
Control architectures for robotic quadrupedal locomotion have seen various
forms. One of the common approaches leverages mathematical optimization
techniques~\cite{chow1971studies} to generate reference trajectories by
solving an optimal control (OC)~\cite{bryson2018applied} problem with
objectives such as minimization of energy consumption, and constraints that
consider the dynamics of the robotic systems. Authors
of~\cite{winkler2018gait} presented a TO formulation for legged locomotion
that automatically generates reference motions without requiring any prior
footstep planning.

Extending upon OC, some of the work has focused on formulating locomotion as
multiple tasks~\cite{bellicoso2016perception}, such as maintaining robot
stability and tracking desired limb motions, solved by prioritizing each
individual task using quadratic programming~\cite{frank1956algorithm} solvers.
This principle of sub-dividing tasks into simpler problems has also been
followed in some of the deep RL research~\cite{lee2019robust}.

Model-based control methods often use
simplified mechanical models to ease on the mathematical complexity of the
problem. For example, most formulations consider the robotic system as a point
mass with massless limbs. These approximations result in control solutions that
cannot exploit the full range of capabilities of the systems. Moreover,
several model-based controllers require hand-tuned costs by human-experts
which are specific to each system's task thereby limiting their generality.

Deep RL methods attempt to address some of these limitations of model-based
control by employing model-free techniques which optimize over a control policy,
a neural network which maps states into actions, so as to maximize a
task-specific reward signal by means of trial and error. Such methods have
been investigated for legged locomotion tasks
\cite{peng2017deeploco,nagabandi2018neural}
but have mainly been demonstrated in simulations using unrealistic robot
models, e.g. ideal torque sources, infinite velocity/torque ranges. Moreover,
RL techniques usually require large amounts of data making training on a real
robot infeasible. This necessitates the use of physics simulators.
However, policies trained using simulations often do not
transfer for real world tasks since the reality gap between simulations and
the physical world are strongly pertinent. These issues of RL have been tackled
using techniques such as actuator modeling~\cite{hwangbo2019learning},
implementing a modular training approach~\cite{zhang2016modular}, introducing domain randomization~\cite{tobin2017domain},
increasing policy generalization~\cite{cobbe2018quantifying} and adding
noise to observations and actions in the training environments. Authors
of~\cite{hwangbo2019learning} have demonstrated the use of a deep RL approach
to complex legged locomotion tasks. Our work, extends upon their training
methods to a constrained learning approach as discussed in
section~\ref{training}.

Despite the success of deep RL approaches on real-world robotic applications,
one of the main challenges in solving an RL problem is precise reward function
tuning. As a solution, an inverse reinforcement learning
(IRL)~\cite{ng2000algorithms} problem can be characterized as: given reference
trajectories of an agent in various circumstances, determine the reward
function to be minimized. This reward function is then used to solve an RL
problem. The authors of~\cite{wulfmeier2017large} and~\cite{finn2016guided}
have successfully implemented IRL methods for perception and control tasks,
however, the need for the extra step of solving an RL problem adds to training
delays. Instead, designing the problem as that of behavioral cloning
(BC)~\cite{torabi2018behavioral} gets rid of the reward recovery step, and directly
optimizes over a policy given reference demonstrations. Along similar lines,
guided policy search (GPS)~\cite{levine2013guided} techniques can be used to
reduce training times by directing policy learning in turn avoiding poor local
optima.

Model-based RL~\cite{doya2002multiple} techniques, which require a knowledge of
system dynamics, have also been proposed as an approach to boost convergence
along desired optima. In the pursuit of making RL methods desirable for use in
safety critical systems, methods such as constrained policy optimization
(CPO)~\cite{achiam2017constrained} have also been investigated to ensure that
an RL control policy obeys the necessary safety constraints during operation.

\subsection{Contributions}
Our work extends upon the above research to realize an RL problem formulation
that considers the constraints required to guarantee the stability of a
quadrupedal robot system. Furthermore, our problem formulation introduces
constraints such as end-effector boundaries, joint velocity limits, and joint
acceleration limits that direct
policy optimization towards a desired quadrupedal
locomotion behavior. This constrained formulation, coupled with techniques
motivated
by BC and GPS, which in our case is guided policy updates (GPUs), further results in the reduction of
training time while also eliminating the need for precise reward function tuning.

We also present the importance of the setup of an RL environment, and
show how differences in dynamic properties can result in a significantly
different learnt behavior. We also compare different physics simulation frameworks and detail upon the motivations of preferring one over the others.

We introduce several schemes that make the quadrupedal system more robust and
therefore better suited for use in real-world applications.
Since we do not use
approximations required for model-based control, our learnt policies better
utilize system dynamics to generate efficient locomotion behavior requiring
significantly lesser
torque compared to a model-based trot controller.
We successfully transfer the control policy trained in a simulator
to the real system and further
provide evidence of dynamic behavior of the control policy by testing
its response after changing the physical properties of the real system, and
also by continuously varying control step times.

\section{Approach}
\label{approach}
In this section we describe the RL methods we use for the task of quadrupedal
locomotion.

\subsection{Algorithm}
Based on the framework of Markov decision processes
(MDPs)~\cite{puterman2014markov}, a constrained Markov decision process
(CMDP)~\cite{altman1999constrained} is defined as a tuple $\left(S, A, R, C, P, d, \mu\right)$, where $S$ is the set of states, $A$ is the set of actions,
$R:S\times A\times A \rightarrow \mathbb{R}$ is the reward function, $P:S\times A\times S\rightarrow \left[0,1 \right]$
is the state transition probablity, and $\mu$ is the starting state
distribution as chracterized in the MDP tuple. CMDPs augment the MDP with a set
$C$ of cost functions, $C_{1},...,C_{m}$, with $C_{i}:S\times A\times
S\rightarrow \mathbb{R}$, and limits $d_{1},\dots,d_{m}$ as described
in~\cite{achiam2017constrained}. Being consistent with the definitions and
notation used by the authors of~\cite{achiam2017constrained}, a stationary
policy $\pi:S\rightarrow \mathcal{P}\left(A\right)$
 is defined as a function mapping states to probability distributions over
actions. The set of stationary policies is defined as $\Pi$.
$\pi\left(a|s\right)$ denotes the probability of selecting action $a$ in state
$s$.

Given a performance measure,
\begin{equation*}
J\left(\pi\right)\doteq\underset{\tau\sim\pi}{\text{\textup{E}}}\left[\sum_{t=0}^{\infty}{{\gamma}^{t}R\left({s}_{t},{a}_{t},{s}_{t+1}\right)}\right]\text{,}
\end{equation*}
where $\gamma\in\left[0,1\right)$ is the discount factor, $\tau$ denotes a
trajectory dependent on $\pi$, we aim to select a policy $\pi$ which maximizes
$J\left(\pi\right)$.
For a CMDP, the expected discount cost return
$J_{C_{i}}\left(\pi\right)\doteq\underset{\tau\sim\pi}{\text{\textup{E}}}\left[\sum_{t=0}^{\infty}{{\gamma}^{t}C_{i}\left({s}_{t},{a}_{t},{s}_{t+1}\right)}\right]$
for a policy $\pi$ with cost function $C_{i}$. The set of feasible stationary
policies $\Pi_{C}\doteq\left\{\pi\in\Pi:\forall_{i},J_{C_{i}}\left(\pi\right)\leq
d_{i}\right\}\text{.}$

The RL problem is then expressed as
\begin{equation*}
  \pi^{\ast}=\argmax_{\pi\in\Pi_{C}}J\left(\pi\right)\text{.}
\end{equation*}

For a policy $\pi_\theta$ parameterized with $\theta$, most policy optimization
strategies iteratively update the base policy using local policy search
methods~\cite{peters2008reinforcement} by maximizing $J\left(\pi\right)$ over a
trust region~\cite{schulman2015trust}. For a CMDP, policy iteration using trust
regions~\cite{achiam2017constrained} can be expressed as
\begin{equation}
  \label{eq:cmdp_policy_iteration}
  \begin{split}
    &\pi_{k+1}=\argmax_{\pi\in\Pi_\theta}\underset{\begin{subarray}{c}{s\sim{d}^{\pi_{k}}}\\{a\sim{\pi}} \end{subarray}}{\text{\textup{E}}}\left[{A}^{\pi_k}\left(s,a\right)\right]\\
    &\text{subject to } J_{C_{i}}\left(\pi_k\right)+\frac{1}{1-\gamma}\left(\underset{\begin{subarray}{c}{s\sim{d}^{\pi_{k}}}\\{a\sim{\pi}} \end{subarray}}{\text{\textup{E}}}\left[{A}_{C_i}^{\pi_k}\left(s,a\right)\right]\right)\leq d_i \quad \forall_i\\
    &\quad\quad\quad\quad \bar{D}_{KL}\left(\pi||\pi_k\right)\leq \delta\text{.}
  \end{split}
\end{equation}

where
$\bar{D}_{KL}=\underset{s\sim\pi_k}{\text{\textup{E}}}\left[D_{KL}\left(\pi||\pi_k\right)\left[s\right]\right]$, and $\delta>0$ is a step size. $D_{KL}$ refers
to the Kullback-Leibler divergence. Authors of~\cite{achiam2017constrained}
show that developing CPO as a trust region method implies CPO inherits the
performance guarantee given by certain lower bound detailed in~\cite{achiam2017constrained}.
The authors of~\cite{achiam2017constrained} also define the worst-case upper bound on the cumulative
discounted return for a CMDP.

The PPO technique introduces a clipped objective function
\begin{equation*}
L_{t}^{CLIP}\left(\theta\right)=\text{\textup{E}}_{t}\left[\text{min}\left(r_t\left(\theta\right)A_t,\text{clip}\left(r_t\left(\theta\right),1-\epsilon,1+\epsilon\right)A_t\right)\right]
\end{equation*}
where $\epsilon$ is a hyperparameter. In our implementation, we introduce an
approximation of the constraint expressed in \eqref{eq:cmdp_policy_iteration}
to the above objective, and rewrite it as
\begin{equation}
  \label{eq:cclip_loss_function}
  \begin{split}
      L_{t}^{CCLIP}\left(\theta\right)=L_{t}^{CLIP}\left(\theta\right) -
\sum_{i}\mathcal{\zeta}_{i}J_{C_{i},t}\left(\pi_{\theta}\right)\text{, }
  \end{split}
\end{equation}
where $\mathcal{\zeta}_{i}$ is an experimentally tuned hyperparameter. The
objective $L^{CLIP}_{t}$ is often augmented to include a value function loss term
$L_{t}^{VF}$ and an entropy term $S$~\cite{schulman2017proximal}. The objective
function, with coefficients $c_{1}$ and $c_{2}$ is then
\begin{equation}
  \label{eq:cppo_loss_function}
  \begin{split}
  &L_{t}^{CCLIP+VF+S}\left(\theta\right)\\
  &=\text{\textup{E}}_{t}\left[L_{t}^{CCLIP}\left(\theta\right) -
c_{1}L_{t}^{VF}\left(\theta\right)+c_{2}S[\pi_{\theta}]\left(s_t\right)\right]\text{.}
  \end{split}
\end{equation}

We introduced this approximation as, in our experiments, we observed that the
constrained policy optimization objective was extremely sample inefficient. We observed no
improvements in our training even after 5 billion sampling steps for the task of tracking base
velocity commands.
As a solution, we implement an approximated constrained proximal policy optimization (aCPPO)
method along
with generalized advantage estimate (GAE)~\cite{schulman2015high}, and optimize
over the loss function represented in~\eqref{eq:cppo_loss_function}. After convergence,
we perform hard constrained proximal policy optimization (hCPPO) using the objective $L_{t}^{CLIP+VF+S}$
and by introducing the cost return constraint expressed in~\eqref{eq:cmdp_policy_iteration}.
We collectively refer to both these optimization steps as constrained proximal policy optimization
(CPPO).

In our work, we introduce three degrees of constraints and handle them
accordingly:
\begin{enumerate}
  \item \textit{Soft} ($\varrho$) constraints are included as part of the
reward function. These need not be critical for safe operations. Instead these
are introduced in order to direct policy search towards a desired behavior.
  \item \textit{Hard} ($\kappa$) constraints cannot be violated and are
included in the set of constrained cost functions $C$. These are directly
included during policy updates. In case of aCPPO, these are included
in~\eqref{eq:cclip_loss_function}, and for hCPPO these are included as a constraint,
as shown in~\eqref{eq:cmdp_policy_iteration} for the objective $L_{t}^{CLIP+VF+S}$.
  \item \textit{No-go} ($\eta$) constraints are introduced during training such
that when $\kappa$-constraints are violated beyond a certain threshold the training
episode is terminated to prevent exploration around regions which do not contribute
towards policy optimization.
\end{enumerate}

We use guided policy updates to warm start our control policy
and then perform constrained proximal policy optimization for policy exploration.
We then alternate between these during training as described in Alg.~\ref{alg:gcpo}.

\begin{algorithm}[htbp!]    % or use \begin{algorithm}[H] for right under
\begin{algorithmic}[1]
\Statex \textbf{Input:} $\lambda_{d}$, $s_{max}$, $c_{max}$, $n_{steps}$,
$n_{batch}$, $in_{user}$, $c_{max}^{h}$, $H_{a}$, $H_{h}$
\Statex \textit{Initialize} $\theta$, $\alpha=1$

\State Generate set of trajectories $\mathcal{D}=\Call{Controller}{in_{user}}$
\Statex \Comment{using model-based control strategy}
\For{$t = 0,1,2\dots H_{a}$}
\State $\theta, \mathcal{D}=\Call{GuidedPolicyUpdate}{\theta, \mathcal{D},
\alpha s_{max},n_{batch}}$
\State $\theta=\Call{PolicyOptimization}{\theta,n_{steps},\left(1-\alpha\right)
c_{max}, false}$
\State $\alpha=\text{\textup{e}}^{\lambda_{d}t}$
\EndFor
\For{$t = 0,1,2\dots H_{h}$}
\State $\theta=\Call{PolicyOptimization}{\theta,n_{steps},
c_{max}^{h}, true}$
\EndFor
\Function{GuidedPolicyUpdate}{$\theta$, $\Tau$, $it_{max}$, $n_{batch}$}
\State $l_{batch} = it_{max} / n_{batch}$
\State Sample $l_{batch}$ state-action pairs $\left(s,a^\ast\right)$ from $\Tau$
\For{$i = 0$ \textbf{to} $n_{batch}-1$}
  \State Generate $\left\{a\right\}$ for $\left\{s\right\}$ using $\pi_\theta$
  \State Update $\theta$ by minimizing $\sum
_{j=0}^{l_{batch}}{||a_{j}-a_{j}^{\ast}||^{2}}$
\EndFor
\State $\Tau = \Tau \setminus \left\{\left(s,a^\ast\right)\right\}$
\State \textbf{return} $\theta$, $\Tau$
\EndFunction

\Function{PolicyOptimization}{$\theta$, $l_{episode}$, $it_{max}$, hard}
\State $it = 0$, $\tau=\{\}$
\While{$it<it_{max}$}
\For{$t = 0$ \textbf{to} $l_{episode}-1$}
  \State Sample $a_t\sim\pi_\theta\left(a|s_{t}\right)$
  \State $s_{t+1},r_{t},\left\{c_{t}\right\}= \Call{RLEnvironmentStep}{a_t}$
  \State $it=it+1$,
$\tau=\tau\cup\left(s_{t+1},r_{t},\left\{c_{t}\right\}\right)$
\EndFor
\If{hard == $false$
    \State Update $\theta$ using aCPPO
\Else
   \State Update $\theta$ using hCPPO
\EndIf}
\EndWhile
\State \textbf{return} $\theta$
\EndFunction

\end{algorithmic}
\caption{\label{alg:gcpo} Guided Constrained Policy Optimization}
\end{algorithm}
\subsection{Simulation}
Most RL algorithms are sample inefficient and require significant amount of
trials to learn a desirable control policy. Instead of training on a physical
platform, which is slow and unsafe, we train the locomotion policy on a
significantly faster simulation environment. However, policies trained in
simulations often do not perform well in real-world systems. This is mostly due
to the reality gap associated with simulations which do not perfectly model the
physical world. Moreover, while performing experiments we realized different
actuator models for ANYmal resulted in considerably different behaviors for the
same training parameters.

We tested ANYmal simulations in RaiSim~\cite{raisim},
PyBullet~\cite{coumans2019} and MuJoCo~\cite{Todorov2012MuJoCoAP} for a simple
task of moving forward with maximum feasible base velocity in order to compare
the generated behaviors and training simulation times. The input to the control
policy (34-dimensional state vector) consisted of $\left\{base_{height},
base_{orientation}, base_{twist}, joint_{states}\right\}$, and the output
(12-dimensional action vector) consisted of $\left\{joint_{positions}\right\}$
desired for the next state. We trained the policies using PPO with the same
hyperparameters, and on the same device, using the reward function $0.3\times base_{forwardVel} -
4\mathrm{e}{-5}\times{\|joint_{torque}\|}^{2}\text{.}$

\begin{figure}[t]
\centering

\begin{subfigure}[b]{0.45\textwidth}
  \centering
  \includegraphics[width=\textwidth]{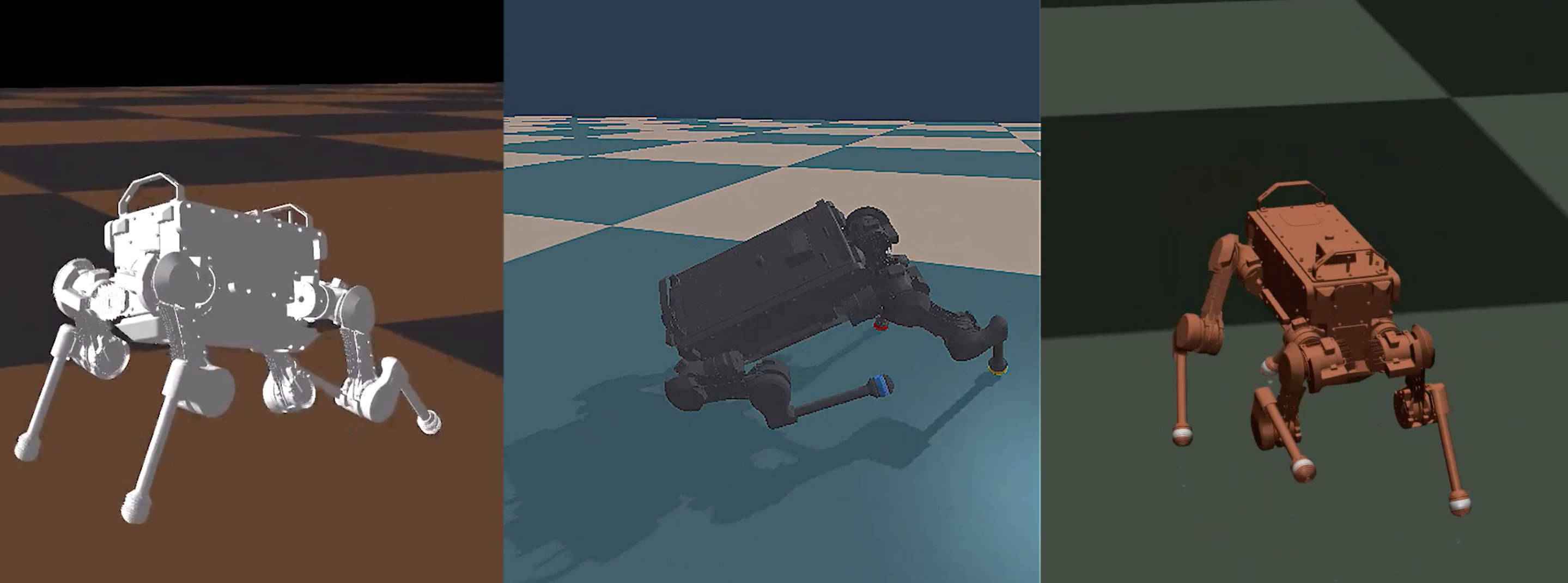}
\end{subfigure}

\vspace{-8pt}
\moveright9.5pt\vbox{\hrule width0.45\textwidth}\nointerlineskip
\vspace{1.5pt}

\begin{subfigure}[b]{0.45\textwidth}
  \centering
  \begin{subfigure}[b]{0.48\textwidth}
    \centering
    \includegraphics[width=\textwidth]{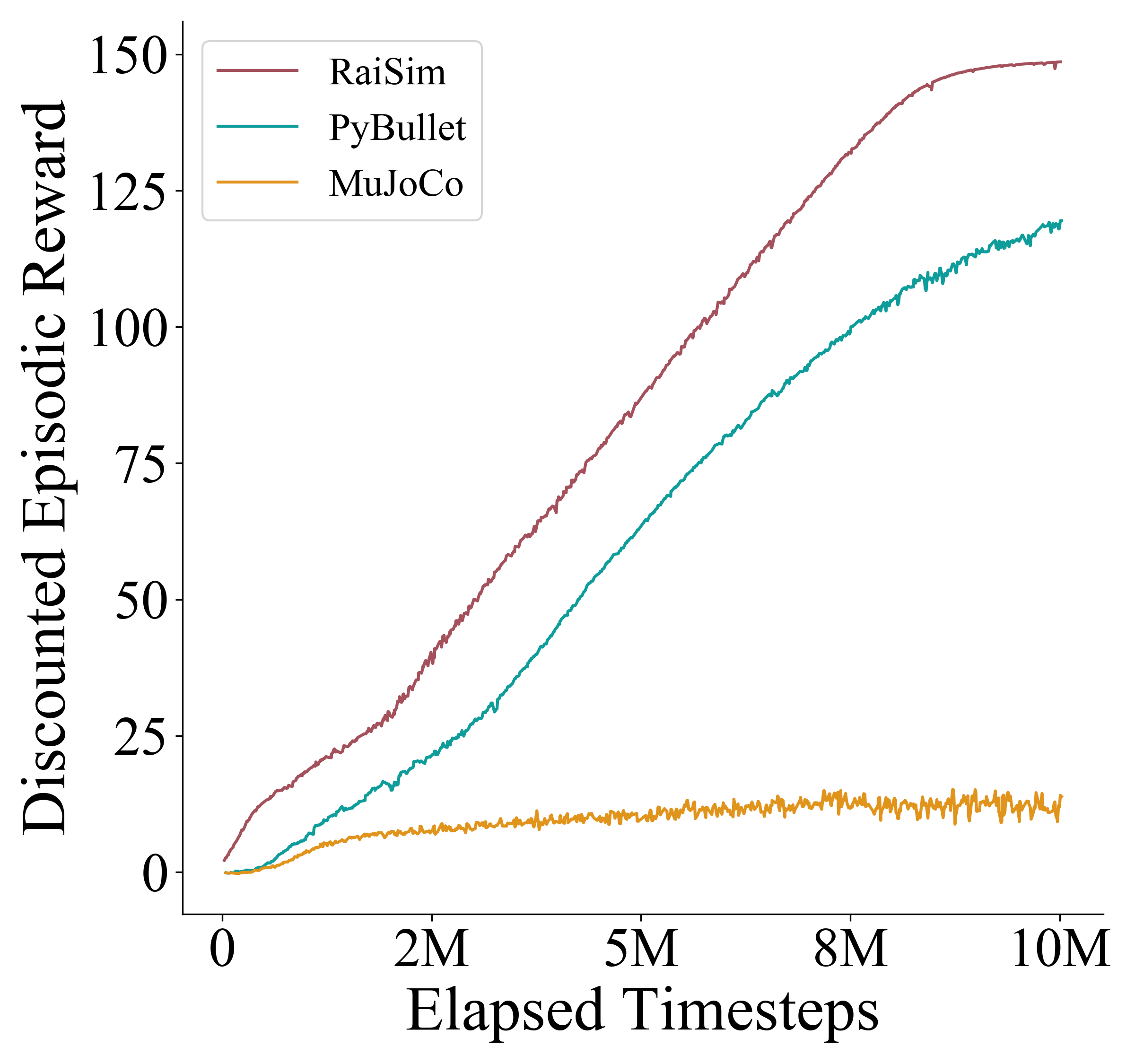}
  \end{subfigure}
  \begin{subfigure}[b]{0.48\textwidth}
    \centering
      \includegraphics[width=\textwidth]{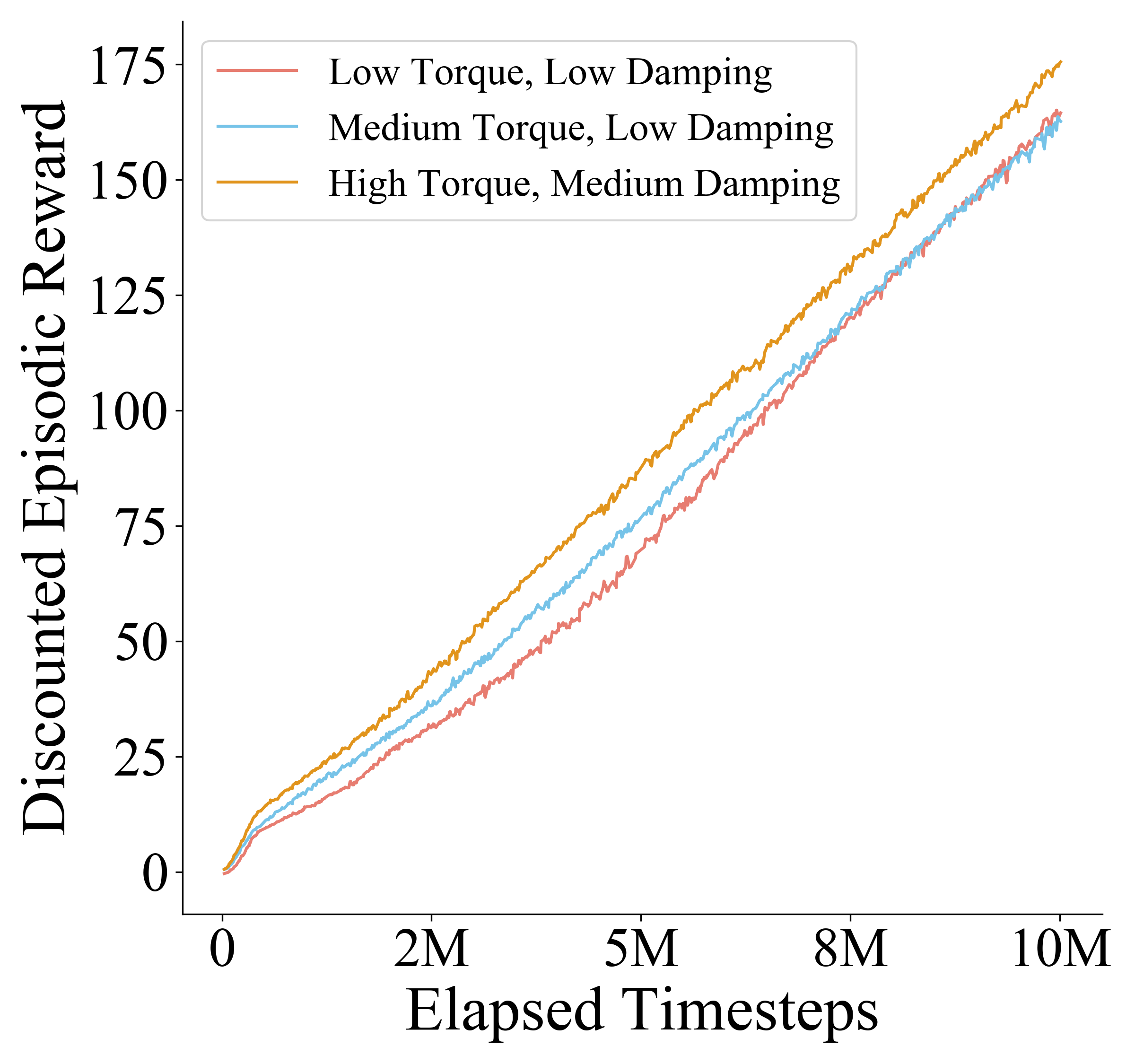}
  \end{subfigure}
\end{subfigure}

\caption{Comparing differences in learnt behavior for different actuator
models. \textbf{(top)} \textit{left:} the RaiSim environment manages to sprint
forward using a gait similar to galloping, \textit{center:} the PyBullet
environment converges at a local optimum where it uses hind legs to push its
body forward while the front legs are not used, \textit{right:} the MuJoCo
environment struggles to find an optimal policy mainly because the desired
joint positions at each step are not tracked reliably. \textbf{(bottom)}
\textit{left:} average discounted reward curves observed during training,
\textit{right:} average discounted reward curves observed during training in
RaiSim for different ANYmal actuator models.}
\label{fig:sim_compare}
\end{figure}

We trained the policies for up to 10M time steps with each iteration comprising
of 76.8k episodic step samples. Using a discount factor $\gamma=0.998$, and
maximum episode length of 6.4k simulation steps we achieved the results
represented in Fig.~\ref{fig:sim_compare}.

We performed experiments using different simulators and actuator
models to validate our point that learnt behavior significantly depends on the
setup of the RL environment, further substantiated by running experiments in
RaiSim using different feed-forward torque and damping parameters for the
actuators, as represented in Fig.~\ref{fig:sim_compare}. Moreover,
none of the policies trained with different actuator models
were observed to be inherently stable, necessitating the use of a good actuator
model. For this, we used the same technique as authors of~\cite{hwangbo2019learning}
to approximate the actuator model using a neural network trained through supervised
learning. We also used the same network architecture as used
in~\cite{hwangbo2019learning} for training the actuator network.

Moreover, due to sample inefficiency in most RL algorithms, it is important to
consider time required for each simulation step. In our experiments, we
observed that for same number of parallel executions, RaiSim was faster than
both PyBullet and MuJoCo as shown in Table~\ref{table:sim_compare_steps}. With
several more parallel executions possible across multiple threads, we managed
to execute 1B simulation steps in less than 3 hours in RaiSim on a PC.

\begin{table}[b]
\caption{Training time required for executing 10M simulation steps using 12
parallel environment runs tested on a PC housing an Intel i7-8700K and an
Nvidia RTX 2080Ti.}
  \begin{center}
    \begin{tabular}{ c | c | c  | c }
      & \textbf{RaiSim} & \textbf{PyBullet} & \textbf{MuJoCo} \\ \hline
      Training Time (seconds) & 1031.6403 & 2043.9825 & 1820.8244 \\
    \end{tabular}
  \end{center}
\label{table:sim_compare_steps}
\end{table}

\subsection{Environment Setup}
\label{subsec:environment_setup}
Authors of~\cite{hwangbo2019learning}~and~\cite{lee2019robust} provide
competitive baselines. We extend their approach to constrained policy
optimization utilizing some of the hyperparameters used in their work. In this
section we describe the setup of the ANYmal RL environment for the task of
tracking user-generated reference base velocity commands.

\subsubsection{Observation Space}
In order to be extendable to the physical robot, the observation space chosen
for the ANYmal environment needs to be accessible through on-board sensors
and state-estimators. In this regard, the 109-dimensional state vector for the
RL environment is defined as $\left\{b_{h}, \mathcal{O}, v_{base},
\omega_{base}, \mathcal{J}_{t}, \mathcal{J}^{des}_{t-1,t-2,t-3,t-4},
\mathcal{\dot{J}}_{t,t-1,t-2}, \mathcal{V}_{base} \right\}$ where $b_{h}$ is
the robot base height, $\mathcal{O}$ is the base orientation, $v_{base}$
 is the linear velocity in base frame, $\omega_{base}$ is the angular velocity
in base frame, $\mathcal{J}_{t}$ is the joint position at time $t$,
$\mathcal{J}^{des}_{t}$ is the policy output at time $t$,
$\mathcal{\dot{J}}_{t}$ is the joint velocity at time $t$ and
$\mathcal{V}_{base}$ is the user-generated desired base velocity expressed in
base frame.

\subsubsection{Action Space}
The control policy outputs a 12-dimensional action vector comprising of
$\left\{\mathcal{J}^{des}_{t}\right\}$. The desired joint positions are
forwarded as an input to the approximated actuator network which outputs the
torques for each of the joints for the ANYmal quadruped. These torques, clipped
between $[-35~Nm,35~Nm]$, are then directly applied to the joints.

\begin{table}[b]
\caption{Reward terms for the MDP formulation. Here $K$ refers to the
logistic kernel defined as ${K(x)=({{e^{x}+2+e^{-x}}})^{-1}}$, $\mathcal{V}^{lin}_{base}$ is the desired linear velocity in
base frame, $\tau$ is the joint torque, $\mathcal{V}^{ang}_{base}$ is the
desired angular velocity in base frame, $v^{foot}_{world, t}$ is the foot
velocity in world frame at time $t$, and $\mathcal{O}_{x,y,z}^{base}$ is the
base orientation along the ${x,y,z}$ axes.}
  \begin{center}
    \begin{tabular}{ c | c }
      Term & Expression \\ \hline \rule{0pt}{3ex}
      Linear Velocity & $K\left(v_{base} - \mathcal{V}^{lin}_{base}\right)$ \\
\rule{0pt}{3ex}
      Torque & $\left\|\tau\right\|^{2}$ \\ \rule{0pt}{3ex}
      Angular Velocity & $K\left(\omega_{base} -
\mathcal{V}^{ang}_{base}\right)$ \\ \rule{0pt}{3ex}
      Foot Acceleration & $||v^{foot}_{world, t} - v^{foot}_{world, t-1}||^2$
\\ \rule{0pt}{3ex}
      Foot Slip & $||v^{foot}_{world}||^2$ \\ \rule{0pt}{3ex}
      Smoothness & $||\mathcal{J}_t - \mathcal{J}_{t-1}||^2$ \\ \rule{0pt}{3ex}
      Orientation & $||\mathcal{O}_{x,y,z}^{base} - \{0, 0,
\mathcal{O}_{z}^{base}\}||^2$
    \end{tabular}
  \end{center}
\label{table:reward_terms}
\end{table}

\subsubsection{Network Architecture}
Since our work focuses on the constrained RL formulation, we decided to
use the same network architecture implemented in~\cite{hwangbo2019learning}, which
had been already demonstrated to perform well.

\subsubsection{Reward Terms}
The reward terms are shown in
Table~\ref{table:reward_terms}. These terms are multiplied by coefficients which are scaled
further to increase the difficulty for the RL agent as training progresses.

\subsubsection{$\varrho$-Constraint Costs}
These costs are directly added to the reward function, and are as
shown in Table~\ref{table:soft_constraints_cost}.

\begin{figure}[tb]
  \centering
  \includegraphics[width=0.5\textwidth]{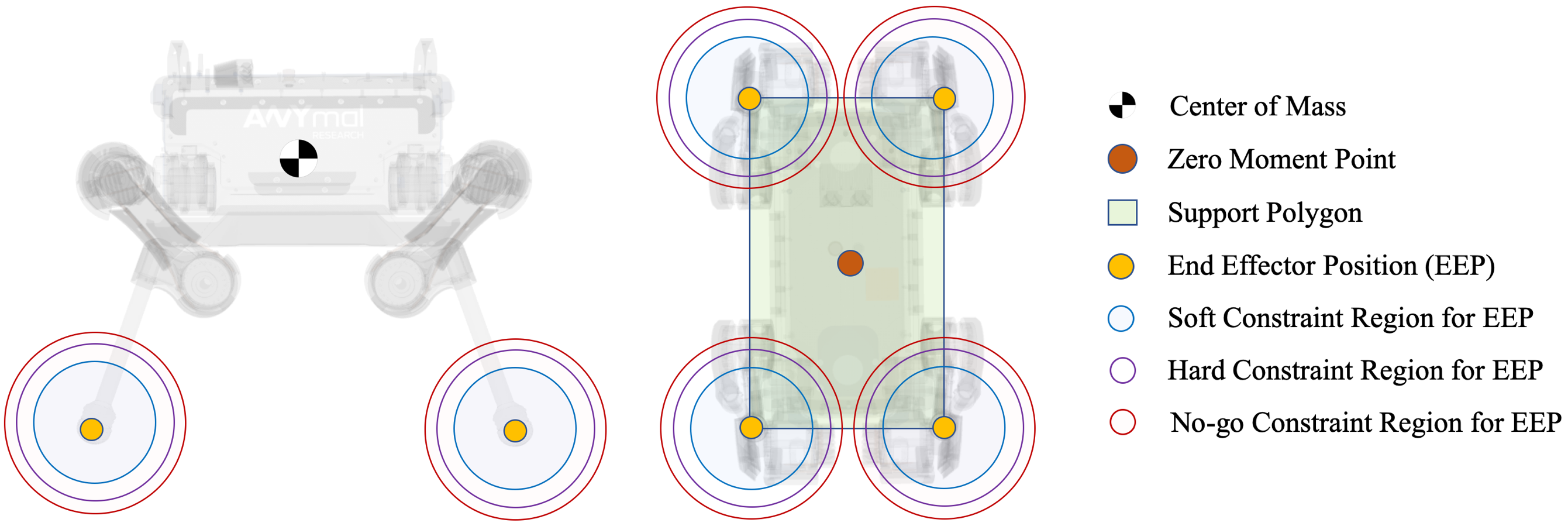}
  \caption{Side-view and top-view of the ANYmal quadruped schematics
representing some of the parameters used for constraint costs.}
  \label{fig:anymal_region}
  \vspace{-0.2 cm}
\end{figure}

\begin{table}[b]
\caption{Cost terms for $\varrho$-constraints. Here
  $\mathcal{\dot{J}}^{limit_{\varrho}}$ refers to the joint speed limit for
  $\varrho$-constraints, $\mathcal{\ddot{J}}^{limit_{\varrho}}$ is the joint acceleration
  limit for $\varrho$-constraints, $f_{h,i}$ is the height of end-effector $i$,
  $f_{h}^{des_{\varrho}}$ is the desired end-effector height, $v_{f,i}$ is the
  velocity of the end-effector $i$, $f_{i}$ is the position of end-effector $i$,
  $\mathcal{R}^{\varrho}_{i}$ is the corresponding feasible end-effector region
  for $\varrho$-constraints, and $f_{i}^{0}$ is the base position for
  end-effector $i$ for a given joint configuration.}
  \begin{center}
    \begin{tabular}{ c | c }
      Term & Expression \\ \hline \rule{0pt}{3ex}
      Joint Speed & $\left\|\text{max}(|\mathcal{\dot{J}}_{t}| - \mathcal{\dot{J}}^{limit_{\varrho}}, 0)\right\|^2$ \\ \rule{0pt}{3ex}
Joint Acceleration & $\left\|\text{max}(|\mathcal{\ddot{J}}_{t}| - \mathcal{\ddot{J}}^{limit_{\varrho}}, 0)\right\|^2$ \\ \rule{0pt}{3ex}
      Foot Clearance & $\sum_{i}\left(f_{h,i} -
f_{h}^{des_{\varrho}}\right)^2\left\|v_{f,i}\right\|^2$ \\ \rule{0pt}{3ex}
      Foot Eligible Region & $\text{bool}\left(f_{i}\notin
\mathcal{R}^{\varrho}_{i}\right)\times \left\|f_{i}^{0}-f_{i}\right\|^2$
\end{tabular}
  \end{center}
\label{table:soft_constraints_cost}
\end{table}

\subsubsection{$\kappa$-Constraint Costs}
These hard constraints are directly introduced in the aCPPO formulation
as part of the expected discount cost return $J_{C_{i}}$ for cost
term $i$. We introduce cost terms that account for stability constraints such as
ZMP, as shown in Fig.~\ref{fig:anymal_region}, that have been used extensively for optimal
control problem formulations.
The $\kappa$-constraint cost terms for the aCPPO are shown in
table~\ref{table:hard_constraints_cost}. We do not use the ZMP term in hCPPO since the violation
of ZMP can be caused due to external perturbations and discarding policies based on ZMP violations
implies that we cannot perform state recovery.

\begin{table}[htbp!]
  \caption{Cost terms for $\kappa$-constraints. Here
  $\mathcal{\dot{J}}^{limit_{\kappa}}$ refers to the joint speed limit for
  $\kappa$-constraints, $\mathcal{\ddot{J}}^{limit_{\kappa}}$ is the joint
  acceleration limit for $\kappa$-constraints, $\mathcal{R}_{i}^{\kappa}$ is the
  corresponding feasible
  end-effector region for $\kappa$-constraints, $u$ is the ZMP, $\mathcal{S}$ is
  the region of support polygon with vertices given by the feet in contact with
  the ground, $\textup{C}$ is the center of mass of the quadruped, and $F_{f_{i}}$
  is the contact force at foot $i$. The foot contacts cost term ensures that
  if 2 feet are in contact with the ground, they are not on the same side.}
  \begin{center}
    \begin{tabular}{ c | c }
      Term & Expression \\ \hline \rule{0pt}{3ex}
      Joint Speed & $\left\|\text{max}(|\mathcal{\dot{J}}_{t}| - \mathcal{\dot{J}}^{limit_{\kappa}}, 0)\right\|^2$ \\ \rule{0pt}{3ex}
Joint Acceleration & $\left\|\text{max}(|\mathcal{\ddot{J}}_{t}| - \mathcal{\ddot{J}}^{limit_{\kappa}}, 0)\right\|^2$ \\ \rule{0pt}{3ex}
      Foot Eligible Region & $\text{bool}\left(f_{i}\notin
\mathcal{R}_{i}^{\kappa}\right)\times \left\|f_{i}^{0}-f_{i}\right\|^2$ \\
\rule{0pt}{3ex}
      ZMP & $\text{bool}\left(u\notin\mathcal{S}\right)\times\left\|u-\textup{C}\right\|^2$
      \\
      \rule{0pt}{3ex}
      Foot Contacts & $\text{bool}(\sum_{i}(F_{f_{i}} > 0)<3\ \& $ \\
      & $((F_{f_{lf}} > 0\ \&\ F_{f_{lh}} > 0) \text{ or } (F_{f_{rf}} > 0\ \&\ F_{f_{rh}} > 0)))$
    \end{tabular}
  \end{center}
\label{table:hard_constraints_cost}
\end{table}

In order to encourage recovery into a stable state upon violation of
$\kappa$-constraints, we introduce an additive reward term for each of the constraints
in case the robot state shifts back to obeying these constraints upon violations.

\subsubsection{$\eta$-Constraint Costs}
For cases when the control policy executes actions that cause the robot to land
in unstable and unrecoverable states, we introduce $\eta$-constraints. Upon
violations of these constraints we terminate the training episode, and disregard
the training steps that may have been explored between $\kappa$-constraint and
$\eta$-constraint violations and add a negative terminal reward to the last
training step in the reformatted episode samples. The intuition behind this was
to limit updates through explorations in regions that, apart from violating
constraints, do not contribute to learning a feasible and desired locomotion
behavior. The constraint terms are shown in table~\ref{table:no_go_constraints_cost}.
When any of the expressions evaluate to true, the training episode is terminated.

\begin{table}[tb]
\caption{$\eta$-constraint terms. Here
  $\mathcal{\dot{J}}^{limit_{\eta}}$ refers to the joint speed limit for
  $\eta$-constraints, $\mathcal{\ddot{J}}^{limit_{\eta}}$ is the joint
  acceleration limit for $\eta$-constraints, $\mathcal{R}_{i}^{\eta}$ is the
  corresponding feasible
  end-effector region for $\eta$-constraints and $u^{\eta}$ is the maximum allowed
  ZMP distance from the center of mass, $\textup{C}$ is the center of mass of the quadruped, and the
  minimum foot contacts have been set to 1 to ensure the control policy does not generate a behavior such as
  pronking.}
  \begin{center}
    \begin{tabular}{ c | c }
      Term & Expression \\ \hline \rule{0pt}{3ex}
      Joint Speed & $\text{bool}(\mathcal{\dot{J}}>\mathcal{\dot{J}}^{limit_{\eta}})$ \\ \rule{0pt}{3ex}
Joint Acceleration & $\text{bool}(\mathcal{\ddot{J}}>\mathcal{\ddot{J}}^{limit_{\eta}})$ \\ \rule{0pt}{3ex}
      Foot Eligible Region & $\text{bool}\left(f_{i}\notin
\mathcal{R}_{i}^{\eta}\right)$ \\ \rule{0pt}{3ex}
      ZMP & $\text{bool}\left(\left\|u-\textup{C}\right\|>u^{\eta}\right)$
      \\
      \rule{0pt}{3ex}
      Foot Contacts & $\text{bool}\left(\sum_{i}\text{bool}(F_{f_{i}} > 0)<2\right)$
    \end{tabular}
  \end{center}
\label{table:no_go_constraints_cost}
\vspace{-0.5 cm}
\end{table}

\section{Training}
\label{training}
An overview of the training and validation process employed for our RL task
is represented in Fig.~\ref{fig:overview}.

\subsection{Generation of Reference Trajectories}
We used a whole-body trotting controller to generate reference trajectories
sampled at 400 Hz using the Gazebo simulator. The trajectories were represented
as state-action $(s,a^\ast)$ pairs as detailed in Section~\ref{subsec:environment_setup}.

\subsection{Guided Policy Updates}
Our training method alternates between supervised learning and reinforcement
learning as presented in Alg.~\ref{alg:gcpo}. The policy updates through
supervised learning ensure that the policy search is directed towards a desired
behavior. We trained the policy using the mean-squared-error loss between
$a^\ast$ and $\pi_{\theta}(s)$ minimized using the Adam~\cite{kingma2014adam}
optimizer.

While performing experiments, we observed that the policy action entropy had to
be reduced after each session of supervised learning, else the policy search
still preferred exploration. In fact, during some training experiments we found
that, without precise reward function tuning, not reducing entropy caused the
RL agent to converge at a local minimum. We empirically determined the
reduction in entropy after each successive update.

\subsection{Constrained Policy Optimization}
During policy exploration and optimization, we use the reward and cost
functions described in the previous section to train the RL agent.
We also introduce the following schemes to make our controller robust to
unaccountable factors.

\subsubsection{Adding Noise to Observations and Actions}
We add Gaussian noise to the state and action vector~\cite{schulman2015trust} to
account for sensor noise and inaccurate actuation. The standard deviation
vector for the
observation space is given as $s_{c}\left\{ 0.02, 0.1, [0.05]_{3}, [0.07]_{3},
[0.02]_{12}, [0.0]_{48}, [0.05]_{36}, [0.0]_{3}\right\}$, and for the action
space, it is given as $s_{c}\left\{[0.04]_{12}\right\}$, where
$s_{c}\in\left[0, 1\right]$
is a scaling term increased over the training period.

\subsubsection{Changing Gravity}
We randomly sample acceleration due to gravity between $\left[0.95g,1.05g\right]$, where $g=9.81$ m/s$^2$
to emulate inertial scaling.

\subsubsection{Actuator Torque Scaling}
We randomly scale the output torque of our actuator network with the scaling coefficient $s_t\in[0.5,2.0]$
to account for differences between the real actuators and the
approximated model.

\subsubsection{Changing Link Mass and Size}
To ensure the training does not converge to a local minimum, we scale the mass and size of
each of the links by coefficients
$s_{m}^{link}\in[0.93,1.07]$, and $s_{l}^{link}\in[0.97,1.05]$ respectively.

\subsubsection{Adding Actuator Damping}
We emulate actuator damping by changing the output of the control policy using a complementary
filter given as $\mathcal{J}_{t}^{des'}=K_{damp}\mathcal{J}_{t}^{des}+(1-K_{damp})\mathcal{J}_{t-
1}^{des'}$ where the gain $K_{damp}$ is randomized between $[1-(s_c/4),
1]$.

\subsubsection{Changing Simulation Step Time}
For the possibility of execution on soft real time systems, we randomly set the step times
of the control loop between
$[2.25, 2.75]$ ms. During experiments we observed that the control policy even
worked when the control frequency was changed from 400Hz to 200Hz.

\section{Results and Discussion}
\label{results_and_discussions}
We performed all training and experiments on commodity hardware; an Intel
i7-8700K and Nvidia RTX 2080Ti. For training a control policy using our
method, we required 450M simulation samples for about 240 policy
optimization iterations using aCPPO, requiring less than 2 hours for a RaiSim based
simulation. We then performed 36 policy iterations over 20M simulation samples using hCPPO.
The authors of~\cite{hwangbo2019learning} required more than
7B simulation steps for convergence.

Having obtained a visually stable behavior in RaiSim, we tested the control policy in Gazebo
simulation of ANYmal which consisted of an analytical actuator model. We then tested the control
policy on the physical system. In our RL training, we do not use Gazebo because the compute time
required for each simulation step is significantly larger than for RaiSim.

\begin{figure}[tb]
\centering
  \includegraphics[width=0.5\textwidth]{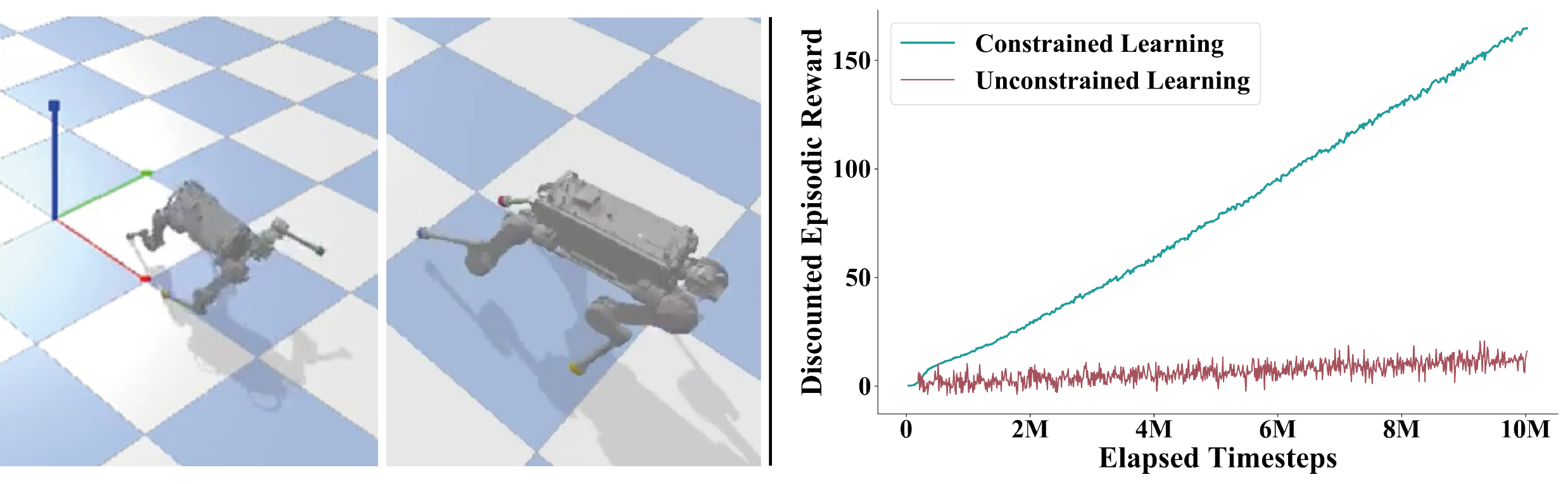}
\caption{Comparing differences in learnt behavior for constrained and
unconstrained learning approaches.
\textit{left:} the unconstrained learning approach fails to develop
an optimal strategy to track the desired
base velocity commands even after 10M steps, \textit{center:} the constrained
learning approach limits the exploration of the policy within desired
regions thereby directing policy optimization towards a preferred optimum,
\textit{right:} the average discounted episodic reward curves observed during
training.}
\vspace{-0.2 cm}
\label{fig:constrained_unconstrained_plot}
\end{figure}

During our experiments with
unconstrained learning methods, we observed that a constrained learning
approach significantly directed policy convergence. We trained a control policy
for a simple task of tracking forward base velocity of 0.7 m/s and observed
that
when we introduced even a basic constraint such as limits on end-effector
positions with respect to the nominal stance, the policy trained using our
constrained proximal policy optimization method performed much better than
unconstrained proximal policy optimization, as is represented in
Fig.~\ref{fig:constrained_unconstrained_plot}. The reward for both the
approaches was defined using the logistic kernel as
$K\left(v_{base} - 0.7\right)$ scaled by a constant. We used the eligible foot
regions defined in
Table~\ref{table:hard_constraints_cost} for constrained learning.

Moreover, for tracking base velocity commands, we required a minimum of
about 2B simulation
samples with precise reward function tuning to obtain a control policy similar
to the trot controller. We changed the reward coefficients,
increasing it for the torque required and decreasing for foot slip, foot clearance and
smoothness empirically over at least 20 trials to get
such a behavior. Without reward tuning, we obtained inefficient locomotion
strategies such as pronking. This was, however, not the case with GCPO. Moreover,
introducing GPUs in our approach helped us reduce the required training samples from
approximately 1.6B, in the case of only CPPO, to 470M for GCPO.

\begin{figure}[tb]
\centering
  \includegraphics[width=0.5\textwidth]{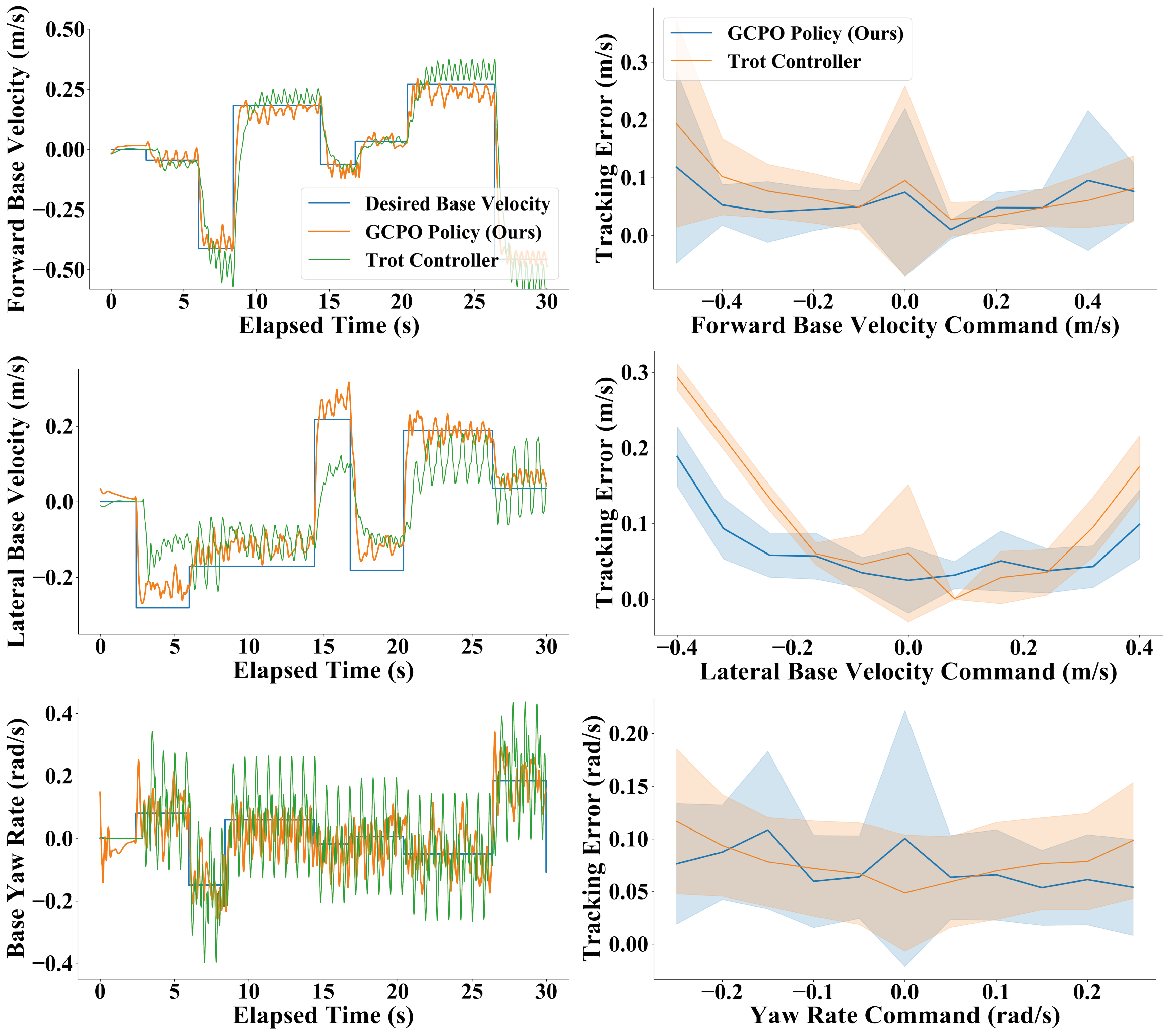}
\caption{Velocity tracking results obtained on the physical quadruped system.
\textit{Left:} base velocities measured for our trained control policy and model-based
trot controller for the same sequence of velocity commands,
\textit{right:} the mean tracking error observed for given base velocity commands. The
confidence bands represent the standard deviation of tracking error.}
\label{fig:base_velocity_tracking}
\end{figure}

\begin{figure}[tb]
\centering
  \includegraphics[width=0.4\textwidth]{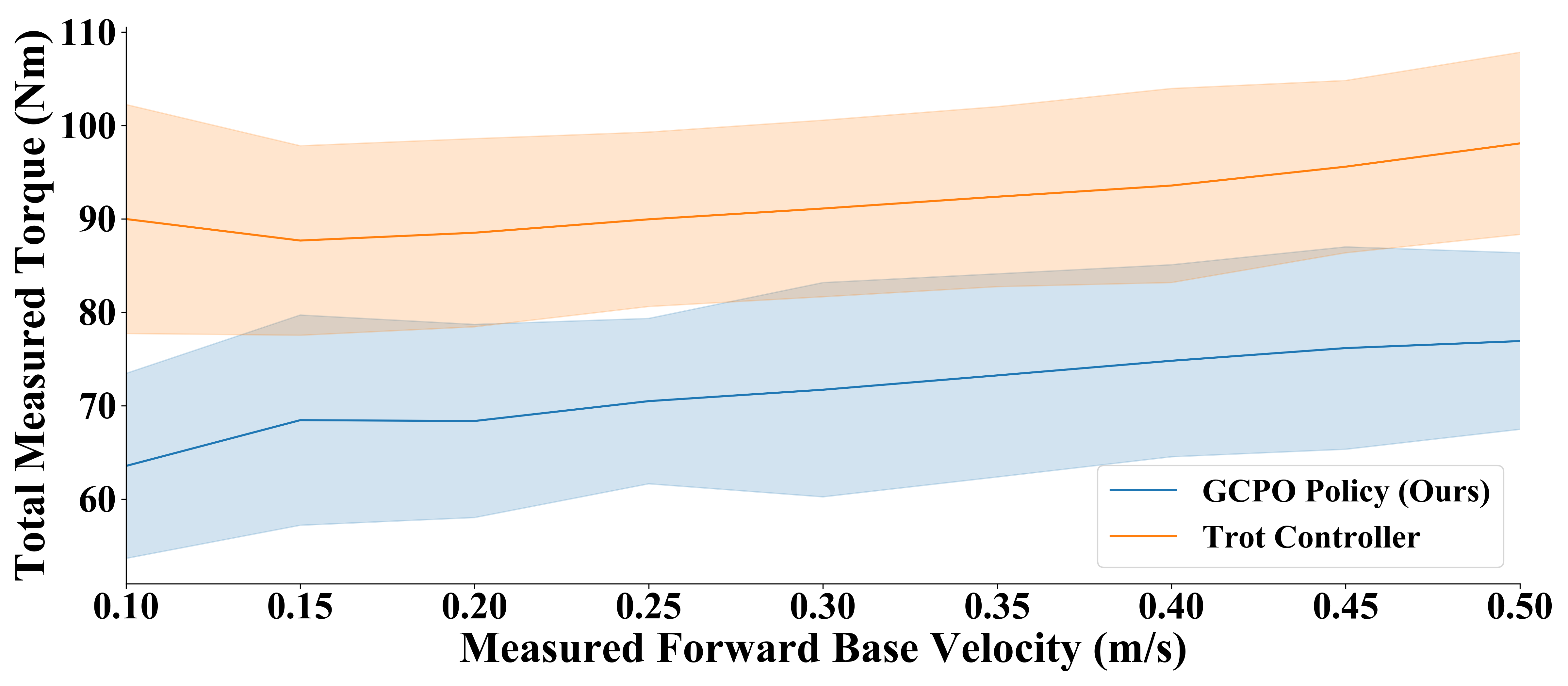}
\caption{The sum of the magnitude of joint torques measured on the physical
system for each of the joints plotted against the measured forward base velocity. The
confidence bands represent the standard deviation of torques measured. The controllers
operate at 400 Hz.}\vspace{-0.1 cm}
\label{fig:torque_compare}
\end{figure}

The velocity tracking results obtained with our trained control policy on
the physical robot system, outdoors on uneven terrain, are as shown in
Fig.~\ref{fig:base_velocity_tracking}. It is important to note that the physical
system comprised of additional sensor modules, amounting to approximately 10\% of the robot
mass, which had not been included in the simulations during training. We compared the
results with the
model-based trot controller we used for GPUs and observed that in most cases our
controller tracks the velocity commands better than the trot controller as is evident
from the tracking error plots. Here, we show that our policy closely tracks the velocity
commands on the physical system despite having been trained on a simulator making this a
successful sim-to-real policy transfer.
Figure~\ref{fig:torque_compare} represents the sum of the magnitude of torques
measured at each joint for measured forward base velocities. We show that our controller,
trained using GCPO, requires significantly lesser torque than the model-based trot
controller. At most of the measured forward base velocities, the total joint torque
measured for our controller is 20 Nm less than for the trot controller.

\begin{figure}[tb]
\centering
  \includegraphics[width=0.5\textwidth]{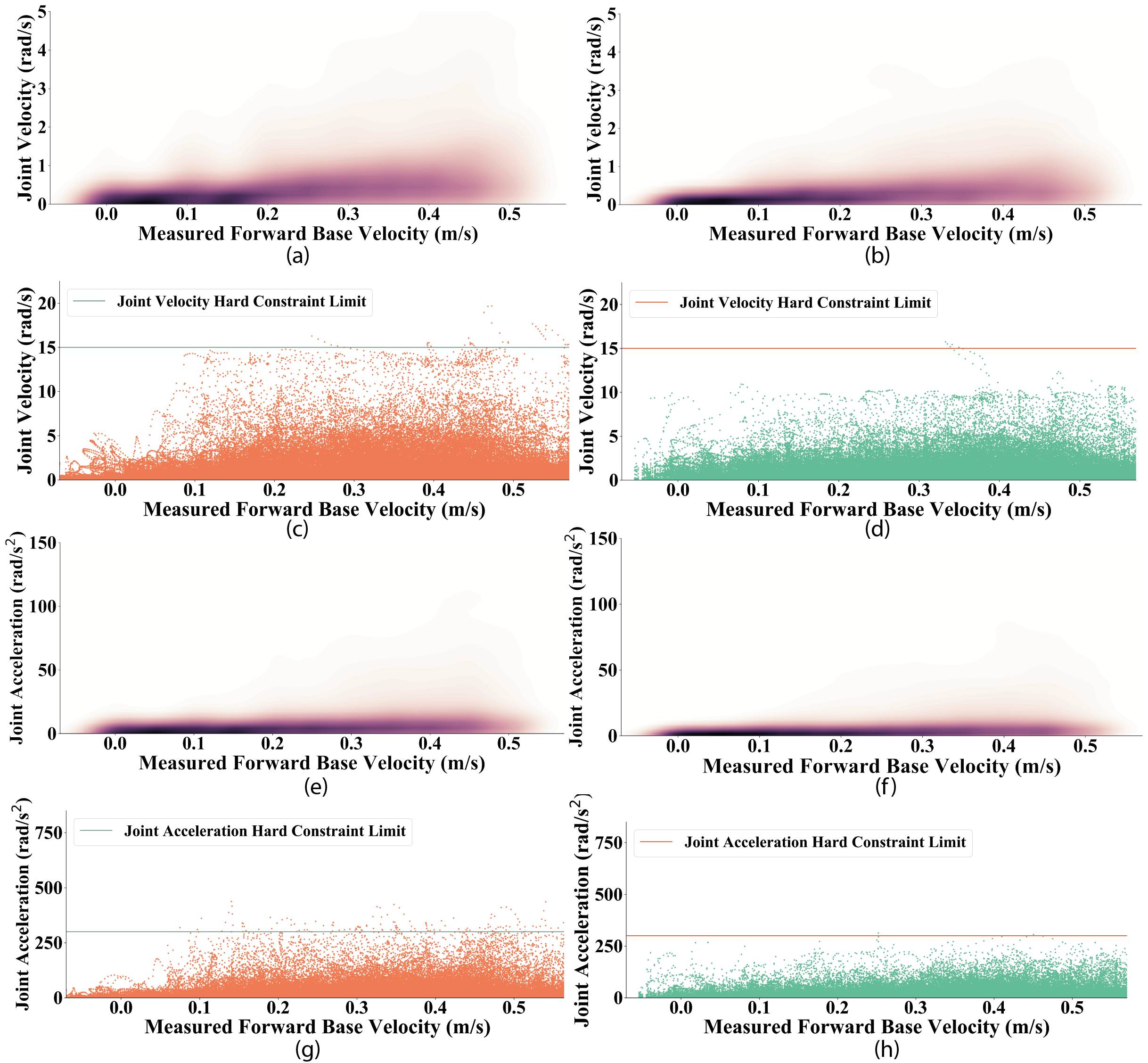}
\caption{Joint velocity and acceleration plots for each of the joint on the ANYmal
quadruped plotted against measured forward base velocity.
\textit{(a)} the kernel density estimate (KDE) for the joint velocities measured on the physical system plotted using
the parameters
detailed in~\cite{scott1979optimal}, \textit{(b)} KDE for joint velocities measured in RaiSim, \textit{(c)} scatter
plot of the measured joint velocities on the physical system with hard constraint limit set to 15 rad/s, \textit{(d)}
scatter plot of
the measured joint
velocities in RaiSim, \textit{(e)} KDE for joint accelerations measured on the physical system, \textit{(f)} KDE
for joint accelerations measured in RaiSim, \textit{(g):} scatter plot of the measured joint accelerations on the
physical system with hard constraint limit set to 300 rad/s$^2$, \textit{(h)} scatter plot of the measured joint
accelerations in RaiSim.}
\label{fig:constraint_following}
\end{figure}

\begin{figure}[tb]
\centering
\vspace{0.1 cm}
  \includegraphics[width=0.5\textwidth]{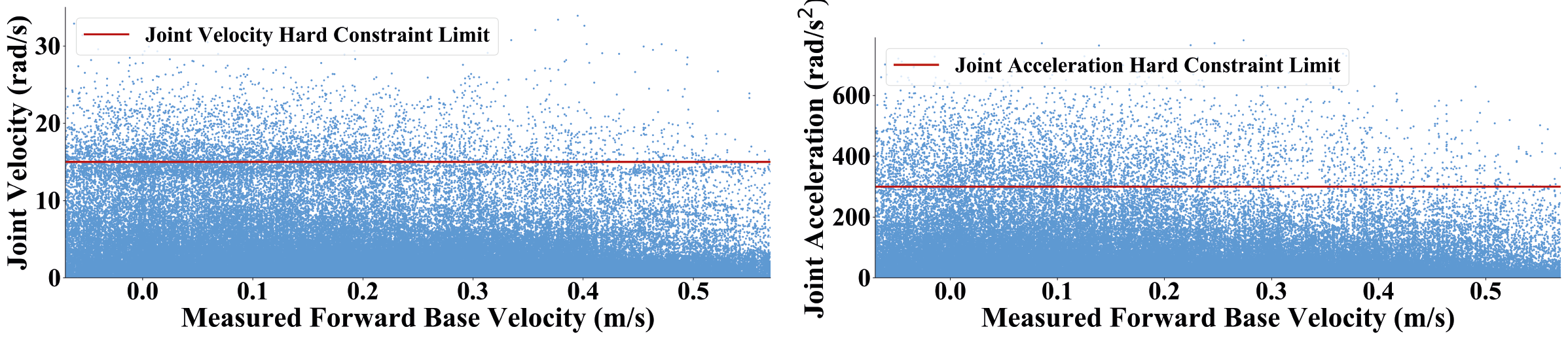}
\caption{
\textit{Left:} Joint velocity scatter plots obtained for every joint on the quadruped for an unconstrained
locomotion behavior measured in RaiSim, \textit{right:} joint acceleration scatter plots.}\vspace{-0.2 cm}
\label{fig:unconstrained_limits}
\end{figure}

%% Constraint following stuff.
As shown in Fig.~\ref{fig:constraint_following}, the density of joint velocity and joint accelerations for
measured forward base velocity commands is very high below 5 rad/s and 50 rad/s$^2$ for the physical robot, while being
even lower for RaiSim. In our experiments, for 5 minutes of data sampled at 400 Hz, we observed that the hard constraints
on the physical system were only violated for 0.0196\% and 0.0490\% of times for joint velocities and joint
accelerations respectively. For RaiSim, these values were 0.00119\% and 0.00178\%. As represented in Fig.~\ref{fig:unconstrained_limits}
for the unconstrained approach, we measured these values in RaiSim and computed them to be
1.1710\% and 1.2511\%, 3 orders of magnitude larger than for our GCPO controller measured in RaiSim. Moreover, the
density of the joint velocities and accelerations is significantly higher near the limits than for our constrained
approach. It is important to note, however, that the behavior of the unconstrained control policies can be changed by significant
re-tuning of the reward function. Furthermore, our control policy always maintains at least 2 stance legs
during locomotion on flat terrain.

Exceeding our expectations, our GCPO controller was able to track the forward base velocity
commands, on the physical system, even
when we introduced delays into the control execution with an RMS tracking error of
0.1736 m/s for following a base velocity command of 0.5 m/s for 10 s. These delays were
randomly sampled from
a uniform distribution between $\left[0, 17.5\right]$ ms. The controller became
unstable when we sampled delays from distributions with upper limit greater than 20 ms.
In most soft-realtime control
systems, these delays are introduced by low-level hardware communication
interfaces, and such a robust controller is certainly desirable.

We observed our controller's response to external perturbations by
applying forces with magnitude ranging from 50 N to 120 N for duration between 1 s
to 5 s to the robot base in Gazebo. We observed that our controller
responded to these external perturbations by moving in the direction opposite to
that of the applied force, ensuring it's stability. The controller was able to respond
to external forces of up to 100 N for 1 s duration applied to the base horizontally.
We also tested the controller's response on the physical system.

We emulated a weak actuator, to test the case
where an actuator becomes damaged, by reducing the
position tracking gain of the weak actuator to 24. The position gain for
all other actuators was set to 48. We observed that our controller was still able
to track the base velocity commands with an RMS tracking error of 0.1975 m/s
for following a base velocity command of 0.5 m/s for 10 s.

Furthermore, despite significant inertial scaling, our control
policy, without any parameter re-tuning, still managed to track velocity commands
even when we set the acceleration due
to gravity to 1.62 m/s$^{2}$ in RaiSim. We observed an RMS tracking error of 0.2214 m/s
for following a base velocity command of 0.5 m/s for 10 s.

\section{Conclusion}
\label{conclusion}
We presented an RL training method for quadrupedal locomotion which
considers safety-critical constraints in its problem formulation and further
encourages system recovery into stable states upon constraint violations. We
used reference trajectories, obtained using a trot controller, to
perform GPUs in order to direct policy optimization. Our
experiments demonstrate that our RL method offers a robust controller which
requires significantly lesser torque for execution compared with a model-based controller.
Furthermore, it is important to note that constrained policy optimization does
not necessitate use of GPUs. In our work, CPPO can be used even without GPUs,
but have been introduced to limit exploration and hence reduce the samples required
for convergence.

As part of future research, we aim to demonstrate the applicability of our
method to more complex environments, and to extend our method to use perception
data as a means to generalizing to a wide variety of challenging terrains.

\section*{Acknowledgement}
\vspace{-0.1 cm}
Siddhant would like to thank Dr. Jemin Hwangbo from RSL, ETH Zurich for his
valuable inputs. We would also
like to thank Luigi Campanaro and Benoit Casseau for helping us with our experiments.

%% Bibliography
\bibliographystyle{IEEEtran}
\bibliography{references.bib}

\end{document}